\pdfoutput=1

\documentclass[11pt]{article}

\usepackage{acl}

\usepackage{times}
\usepackage{latexsym}

\usepackage[T1]{fontenc}

\usepackage[utf8]{inputenc}

\usepackage{microtype}

\usepackage{color}
\usepackage{amsmath}
\usepackage{amsfonts}
\usepackage{amssymb}
\usepackage{booktabs}
\usepackage{graphicx}

\title{---CTI Competition Report (rank 9th) ---\\ 
EMOFM\thanks{\quad \textcolor{cyan}{Emotion is a live radio singing the life.}} : Ensemble MLP mOdel with Feature-based Mixers\\ for Click-Through Rate Prediction}

\author{Yujian Betterest Li$^{\diamondsuit\clubsuit}$\thanks{\quad Correspondence: \url{bebetterest@outlook.com}} 
\quad  
Kai Wu$^{\diamondsuit}$
\\
$^\diamondsuit$School of Artificial Intelligence, Xidian University\\
$^\clubsuit$Institute of Freedom and Happiness, Dream Place\\}

\begin{document}
\maketitle
\begin{abstract}
Track one of CTI competition\footnote{\url{http://cti.baidu.com}} is on click-through rate (CTR) prediction. The dataset contains millions of records and each field-wise feature in a record consists of hashed integers for privacy. For this task, the keys of network-based methods might be type-wise feature extraction and information fusion across different fields. Multi-layer perceptrons (MLPs) are able to extract field feature, but could not efficiently fuse features. Motivated by the natural fusion characteristic of cross attention and the efficiency of transformer-based structures, we propose simple plug-in mixers for field/type-wise feature fusion, and thus construct an field\&type-wise ensemble model, namely \textbf{EMOFM} (\textbf{E}nsemble \textbf{M}LP m\textbf{O}del with \textbf{F}eature-based \textbf{M}ixers). In the experiments, the proposed model is evaluated on the dataset, the optimization process is visualized and ablation studies are explored. It is shown that \textbf{EMOFM} outperforms compared baselines. In the end, we discuss on future work. \textcolor{red}{WARNING: The comparison might not be fair enough since the proposed method is designed for this data in particular while compared methods are not. For example, \textbf{EMOFM} especially takes different types of interactions into consideration while others do not. Anyway, we do hope that the ideas inside our method could help other developers/learners/researchers/thinkers and so on.}
\end{abstract}

\section{Introduction}
\label{sec:intro}

Track one of the CTI competition is on the click-through rate (CTR) prediction task, to predict the probability of whether a certain user would interact with a candidate item in a certain session on real advertisement dataset. The dataset consists of millions of records on 30 continuous days. There are several segments in each record: basic user information $F_{user}$,  scene information $F_{scene}$, advertisement information $F_{ad}$, user session information $F_{session}$, interaction type $F_{type}$ and time stamp $F_{time}$. Each segment is represented by integers of various lengths. Specifically, $|F_{user}|$=13, $|F_{scene}|$=3, $|F_{ad}|$=8, $|F_{session}|$=2, $|F_{type}|$=1 and $|F_{time}|$=1. Besides, each feature is hashed except for $F_{type}$ and $F_{time}$. In addition, click-through label $Y$ is 1 or 0, where 1 stands for that the user does interact with the advertisement in the record. Although interaction types may be different, the goal of the task is to tell the probability of type-agnostic interaction. Besides, it is worth mentioning that $|F_{type}|$ is not directly available for inference.

Since CTR (\citealp{zhang2021deep}; \citealp{YANG2022102853}) is really important for online advertising and recommendation systems, many researchers focus on it. Though basic logistic regression could be easily applied to CTR tasks, considering the characteristic of discrete and sparse feature, there exists many limitations. Focus on the limitations, based on experience of experts or pre-defined formulas, early methods usually pay attention to designing manual structures for effective feature fusion. For example, Factorization Machines (FM) (\citealp{5694074}) utilizes the inner-product of embedding vectors to realize second-order feature combination. With deep learning becoming popular for the efficiency of data-driven optimization and the conciseness of end-to-end structures, deep learning based methods are applied into CTR tasks. However, pure Multi-Layer Perceptrons (MLPs) are weak in implicitly learning feature fusion within a limited scale (\citealp{10.1145/3539618.3591988}). Thus, many works focus on combining explicit feature fusion and implicit data-driven optimization. For example, DeepFM (\citealp{10.5555/3172077.3172127}) introduces Factorization Machines to deep neural networks. In recent years, more and more designed deep learning based components for feature fusion are proposed: DCNv2 (\citealp{10.1145/3442381.3450078}) introduces cross layers to combine hidden features with embedding features; DESTINE (\citealp{10.1145/3459637.3482088}) comes up with disentangled self-attention mechanism for valid fusion; EDCN (\citealp{10.1145/3459637.3481915}) utilizes bridge modules to share information across fields while regulation modules to differentiate feature; FINAL (\citealp{10.1145/3539618.3591988}) extends linear layers to factorized interaction layers for high-order feature fusion.

In this paper, motivated by effective fusion of cross attention and efficient feature extraction of transformer-based structures (\citealp{NIPS2017_3f5ee243}; \citealp{LIN2022111}), we propose plug-in mixers for field/type-wise feature fusion and then construct a field\&type-wise ensemble model \textbf{EMOFM}. Through evaluation, the proposed model shows better performance. We further discover the optimization process and explore ablation studies. In addition, limitations and potential directions for future work are discussed. 

\section{Methodology}

\subsection{Problem Formulation}

As mentioned in Section~\ref{sec:intro}, given feature $F$ and label $Y$, the objective for optimization could be formulated as
$$
\mathop{\arg\min}\limits_{\theta}\sum_{i \in |dataset|}\emph{loss}(M_{\theta}(F_i),Y_i)
$$
where Logloss is considered as \emph{loss} in this paper, $M$ is the prediction model, $\theta$ is learnable parameters of $M$, $F$ consists of $F_{user}$, $F_{scene}$, $F_{ad}$, $F_{session}$, $F_{type}$ and $F_{time}$.

\subsection{Feature-based Mixers}
\label{sec:mixer}

Motivated by the natural capability of feature fusion of cross attention, we attempt to propose a plug-in feature fusion mixer block for various features with various dimensions of various fields. Given two target feature embeddings $E_1\in\mathbb{R}^{d_1}$ and $E_2\in\mathbb{R}^{d_2}$, firstly, they are projected to sequence features as
$$
E_{s,i}=unsqueeze(E_i) W_{p,i}\quad(i=1,2)
$$
where $unsqueeze$ inserts a dimension of size one at the last index, $W_{p,i}\in\mathbb{R}^{1 \times d_h}$, $d_h$ is the dimension of hidden features, and $E_{s,i}\in\mathbb{R}^{d_i\times d_h}$. Besides, $E_{s,i}$ could be considered as sequence feature with the length of $d_i$ and the dimension of $d_h$. Next, we apply cross attention for feature fusion as 
\begin{gather*}
E_{f,1}=cross_{att}(E_{s,1},E_{s,2})\\
E_{f,2}=cross_{att}(E_{s,2},E_{s,1})
\end{gather*}
where
\begin{gather*}
\small
\begin{cases}
cross_{att}(E_{s,i},E_{s,j})=cat(head_1,...,head_{hd}) W^O\\
head_k=Softmax(\frac{(E_{s,i} W^Q_i)(E_{s,j} W^K_i)^\top}{\sqrt{d_{hd}}}) (E_{s,j} W^V_i)
\end{cases}
\end{gather*}
$cat$ refers to concatenate, $hd$ is the number of the heads for cross attention, $W^O\in\mathbb{R}^{d_h\times d_h}$, $W^Q\in\mathbb{R}^{d_h\times d_{hd}}$, $W^K\in\mathbb{R}^{d_h\times d_{hd}}$, $W^V\in\mathbb{R}^{d_h\times d_{hd}}$, and $d_{hd}=d_h/hd$. Respectively, $E_{f,1}$ ($E_{f,2}$) is the fused feature by weighted-aggregation of $E_{s,2}$ ($E_{s,1}$) relating to $E_{s,1}$ ($E_{s,2}$). Then, we apply reversely projection and residual connection as follows
$$
E_i'=E_i+squeeze(E_{f,i} W_{rp,i})
$$
where $squeeze$ removes dimensions of size one. $W_{rp,i}\in\mathbb{R}^{d_h\times 1}$ and $E_i'\in\mathbb{R}^{d_i}$. In this way, we could fuse a pair of features keeping the dimension fixed. The overall process is formulated as 
$$
E_1',E_2'=Mixer(E_1,E_2)
$$

\subsection{EMOFM}
\textbf{EMOFM} is based on a typical deep CTR model paradigm, "Embedding + MLP". For each field, we embed the feature as 
$$
E=cat(flatten(F_1 W^E),...,flatten(F_{|F|} W^E))
$$
where $flatten$ refers to reshaping the dimension into one, $d_e$ is the dimension of embedding, $W^E\in\mathbb{R}^{1\times d_e}$ and the embedding $E\in\mathbb{R}^{|F|d_e}$. $W^E$ is shared to limit the size of the model. Besides, each MLP block consists of a layernorm, a linear layer and a SiLU/Sigmoid activation. The activation is Sigmoid if it is the last black, otherwise SiLU. Next, we construct 3 base models (Whole MLPs (\textbf{WM}), Hierarchical MLPs (\textbf{HM}) and Hierarchical MLPs with Mixers (\textbf{HMM})) step by step for balancing explicit and implicit feature extraction and fusion as follows.
\paragraph{A) WM} Each single model consists of a bottleneck stack of MLP blocks which is applied to concatenated embedding feature of all fields. To strengthen the connection of different $F_{type}$ and the specific patterns of each $F_{type}$, four single models are implemented. One single model, namely uni-predictor, fits every record while another three single models, namely type-wise predictors, would fit the records of the corresponding type. Single models share embedding for consideration of memory and scale. When training, sum of losses of uni-predictor and the corresponding type-wise predictor is considered as the total loss; for inference, average of outputs of uni-predictor and the corresponding type-wise predictor is considered as the final prediction.
\paragraph{B) HM} Based on \textbf{WM}, each single model is split into 2 stages. First, for each field, the feature is independently encoded by the corresponding bottleneck MLP blocks. Since $F_{type}$ and $F_{time}$ are much clear and distinguishable, they are excluded at the first stage for simplicity. Second, another bottleneck stack predicts the probability by all encoded features. In addition, we enrich low-dimension feature by concatenating the projection of embedding features at the second stage. Hence, we manually push the model to firstly pay attention to the field-wise features individually at stage one.
\paragraph{C) HMM} Based on \textbf{HM}, we insert the proposed mixers in front of each MLP block. Specifically, at the first stage, to fuse with different fields, $cat(F_{user},F_{session})$ and $cat(F_{scene},F_{ad})$ are considered as the target features; at the second stage, considering the connection of uni-predictor and type-wise predictors, the hidden states of uni-predictor and the corresponding type-wise predictor are considered as the target features. In this way, we insert explicit field/type-wise learnable mixers at the corresponding stage, leading to a transformer-style structure with staggered modules for extraction and fusion.
\paragraph{Auxiliary Model (AM)} To solve the problem that $F_{type}$ is only accessible when training, we additionally train an auxiliary model for interaction type prediction. The structure of the model is similar to that of \textbf{HMM}. The difference is that input features of \textbf{AM} do not include $F_{type}$, which is the label instead. For scale and memory consideration, embedding is shared from the pre-trained prediction model and frozen when training. Totally, \textbf{AM} is trained after the training of the prediction model; for inference, \textbf{AM} is firstly utilized to predict $F_{type}$ for the prediction model.

Totally, \textbf{EMOFM} is an ensemble of \textbf{WM}, \textbf{HM} and \textbf{HMM} by average aggregation. In a word, we attempt to utilize features of different fields and insert explicit operation structures into a completely implicit model to hint the learning process. All concise illustrations of the introduced structures and methods are shown in Appendix \ref{app-ill}.

\section{Experiments}

\paragraph{Setups} The proposed models are evaluated on the given dataset. Records on the first 29 days are considered as training data $D_{train}$ while the remaining records are considered as test data $D_{test}$. AUC and Logloss are considered as the metrics. For \textbf{WN}, there are 5 MLP blocks with output dimensions of 512, 256, 128, 64 and 1. For \textbf{HM}, there are 2 MLP blocks for each field at the first stage. Specifically, the output dimensions of MLP blocks for basic user information are 128 and 64; those for scene information are 20 and 16; those for advertisement information are 256 and 128; those for session information are 84 and 32. At the second stage, another 2 MLP blocks, with output dimensions of 128 and 1, are included. The dimension of the additional projected feature is 256. For \textbf{HHM}, $d_h$=4, $hd$=2, and a field/type-wise Mixer is inserted in front of each MLP block at the corresponding first/second stage. For all proposed models, $d_e$ of $F_{time}$ and $F_{type}$ is 24 while that of other fields is 8; $F_{time}$ for inference is set to the time stamp of the last day for training; the batch size is 4096 for \textbf{WN} and \textbf{HM}, 1024 for \textbf{HHM}, and 8192 for \textbf{AM}; \textbf{HM} and \textbf{HHM} are respectively ensembled to avoid fluctuation by simply averaging two models with different seeds. In addition, four related methods (DCNv2 (\citealp{10.1145/3442381.3450078}), DESTINE (\citealp{10.1145/3459637.3482088}), EDCN (\citealp{10.1145/3459637.3481915}) and FINAL (\citealp{10.1145/3539618.3591988})) are included for comparison. The batch size of them is 4096; the dimension of embedding is set to 8 as well; the number of hidden units is 4 and the all dimensions are 400; for DCNv2, the parallel structure is considered, the number of experts is set to 4, 4 cross layers are included; for DESTINE, 4 attention layers are included with 4 heads and the attention dimension of 8; for EDCN, 4 cross layers are included, and concatenation is the bridge type; for FINAL, 2B structure is considered. For all models, due to the one-epoch overfitting phenomenon (\citealp{10.1145/3511808.3557479}), we only optimize the models for one epoch. Adam (\citealp{kingma2014adam}) is applied, global learning rate is set to 0.001. For the proposed models, the learning rate of embedding is 0.1 for quick convergence. The rest keeps the default configure. Besides, the proposed models are implemented by PaddlePaddle\footnote{\url{https://github.com/paddlepaddle/paddle}} while other models are implemented by FuxiCTR\footnote{\url{https://github.com/xue-pai/FuxiCTR}} toolkit (\citealp{10.1145/3459637.3482486}). All experiments could run with one V100-32GB GPU.
\begin{table}[ht]
\centering
\small
\begin{tabular}{ccc}
\toprule
Model    & AUC    & Logloss \\ \midrule
baseline & 0.7266 & 0.2192 \\ \midrule
DCNv2    & 0.7405 & 0.2163 \\
DESTINE  & 0.7397 & 0.2166 \\
EDCN     & 0.7411 & 0.2163 \\
FINAL    & 0.7408 & 0.2163 \\ \midrule
\textbf{WM}       & 0.7437 & 0.2155 \\
\textbf{HM}       & 0.7457 & 0.2150 \\
\textbf{HMM}      & 0.7452 & 0.2152 \\
\textbf{EMOFM}    & \textbf{0.7472} & \textbf{0.2147} \\
\bottomrule  
\end{tabular}
\caption{Performance of different models on $D_{test}$. We repeat each experiment twice with different seeds (2000927 and 2000928), and average the values. Standard deviations are not included since they are too small.}
\label{tab:overall}
\end{table}
\paragraph{Overall View}
As shown in Table~\ref{tab:overall}, the proposed methods are better than all compared methods on the dataset. \textbf{HM} and \textbf{HMM} are better than \textbf{WH}, which demonstrates that correct explicit structures would enhance feature fusion. It is strange that \textbf{HMM} is slightly worse than \textbf{HM}, which would be further demonstrated in Ablation Studies. In addition, \textbf{EMOFM} achieves the best AUC and Logloss.
\begin{figure}[ht]
\centering
\small
\includegraphics[width=1.0\columnwidth]{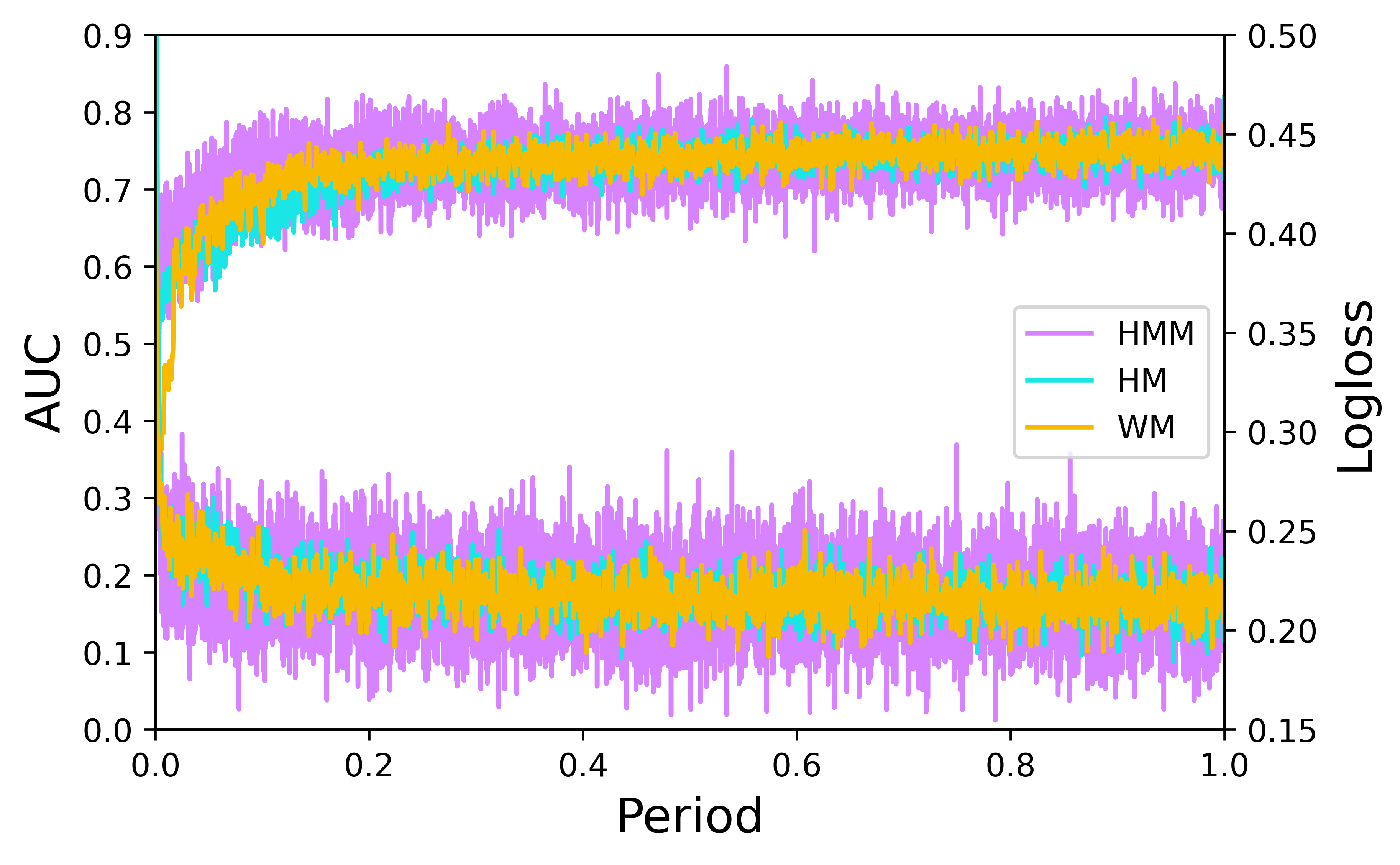}
\caption{AUC\&Logloss vs. the training period on $D_{train}$. The increasing curves describe AUC while the others describe Logloss.}
\label{fig:process}
\end{figure}
\paragraph{Optimization Process} 
As shown in Figure \ref{fig:process}, we discover the optimization process. Since the epoch is set to one, the performance curves on $D_{train}$ could reflect the relatively objective trend of performance. During the training period, AUC increases while Logloss decreases. The curves exhibit rapid changes during the initial 20\% of the training period, gradually followed by a gentler trend in the remaining 80\%.
\begin{table}[ht]
\centering
\small
\begin{tabular}{cccc}
\toprule
Model & base   & w/o type-wise & w/o $F_{type}$ \\ 
\midrule
\textbf{WM}    & 0.7437 & 0.7420        & 0.7429      \\
\textbf{HM}    & 0.7457 & 0.7446        & 0.7445      \\
\textbf{HMM}   & 0.7452 & 0.7445        & 0.7456      \\
\textbf{EMOFM} & 0.7472 & 0.7467        & 0.7466      \\ 
\bottomrule  
\end{tabular}
\caption{AUC performance of ablation studies. w/o type-wise refers to a single uni-predictor while w/o $F_{type}$ refers to excluding interaction type information.}
\label{tab:ablation}
\end{table}
\paragraph{Ablation Studies}
We here attempt to figure out the impact of  ensemble structures and  additional type information $F_{type}$. As shown in Table \ref{tab:ablation}, both type-wise predictors and $F_{type}$ are key factors for the proposed models. Since \textbf{WM} includes less explicit learning patterns, it is much worse than \textbf{HM} and \textbf{HMM} without additional structures or information. In addition, It is worth noting that \textbf{HMM} performs well when $F_{type}$ is not available. Considering \textbf{HMM} is a little worse than \textbf{HM} in Overall View, it is supposed that the balance of feature fusion and field-wise separate extraction for the given data is essential. With more distinguishable information, field-wise separate extraction is much helpful while fusion would pollute the independent field-wise feature, leading to poor performance; when information is relatively vague, field-wise separate extraction could not work well while feature fusion enriches feature representation.

\section{Discussion}

In this work, we apply cross attention to feature fusion and construct an ensemble MLP model with feature-based mixers (\textbf{EMOFM}). The proposed method is better than other models on the given data. Moreover, we explore optimization process and illustrate ablation studies in detail. We find that the balance of implicit and explicit learning patterns and the balance of field-wise separate extraction and feature fusion are important, which may demonstrate that completely data-driven methods have difficulty in feature representation for CTR tasks. We are interested in how to construct a universal data-driven structure which would automatically learn how to handle the balances mentioned above in future work. 

\bibliography{custom}

\appendix

\section{Concise Illustrations}
\label{app-ill}

\begin{figure*}[ht]
\centering
\small
\includegraphics[width=2.0\columnwidth]{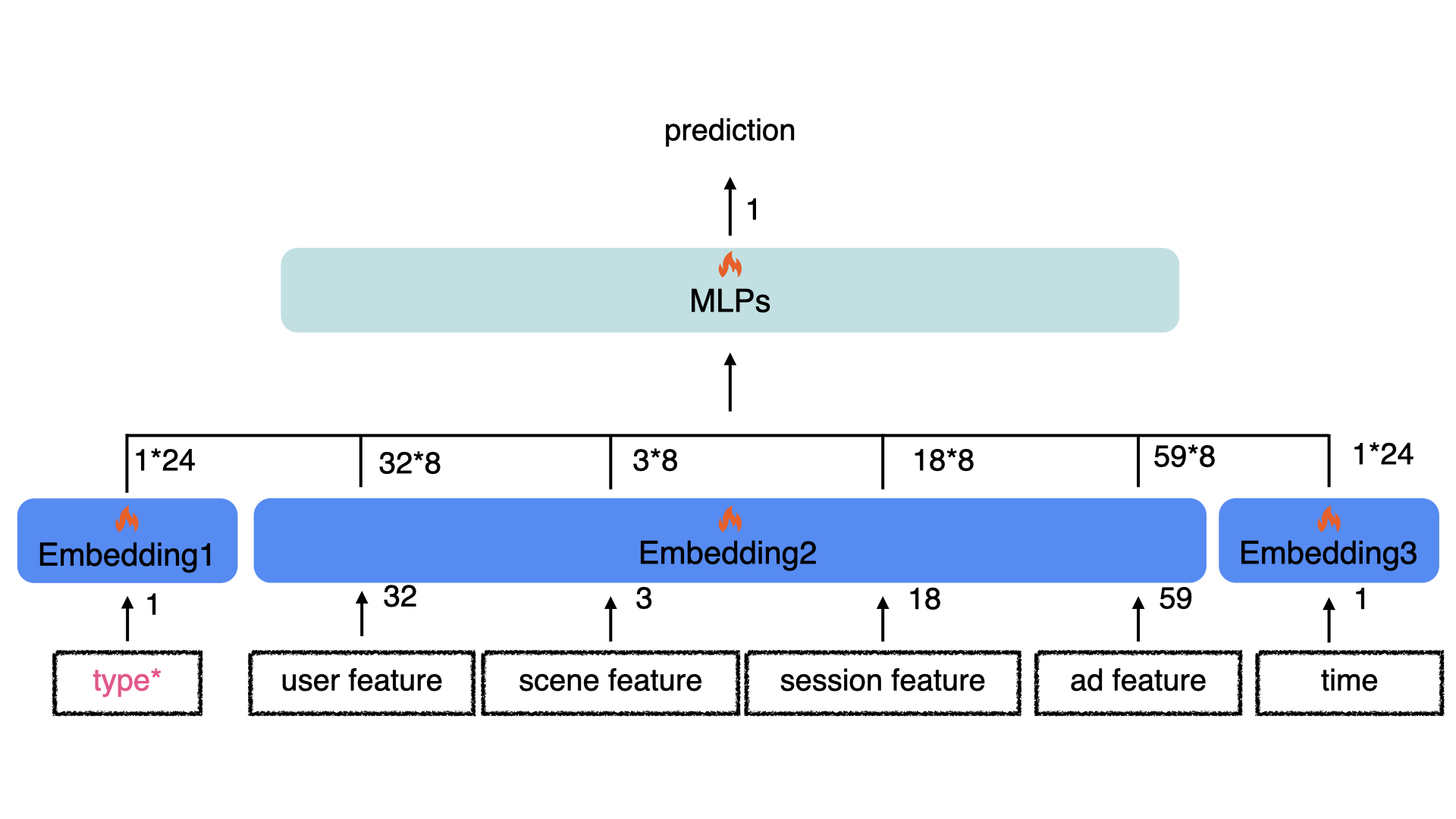}
\caption{Illustration on the structure of Whole MLPs (\textbf{WM}).}
\end{figure*}

\begin{figure*}[ht]
\centering
\small
\includegraphics[width=2.0\columnwidth]{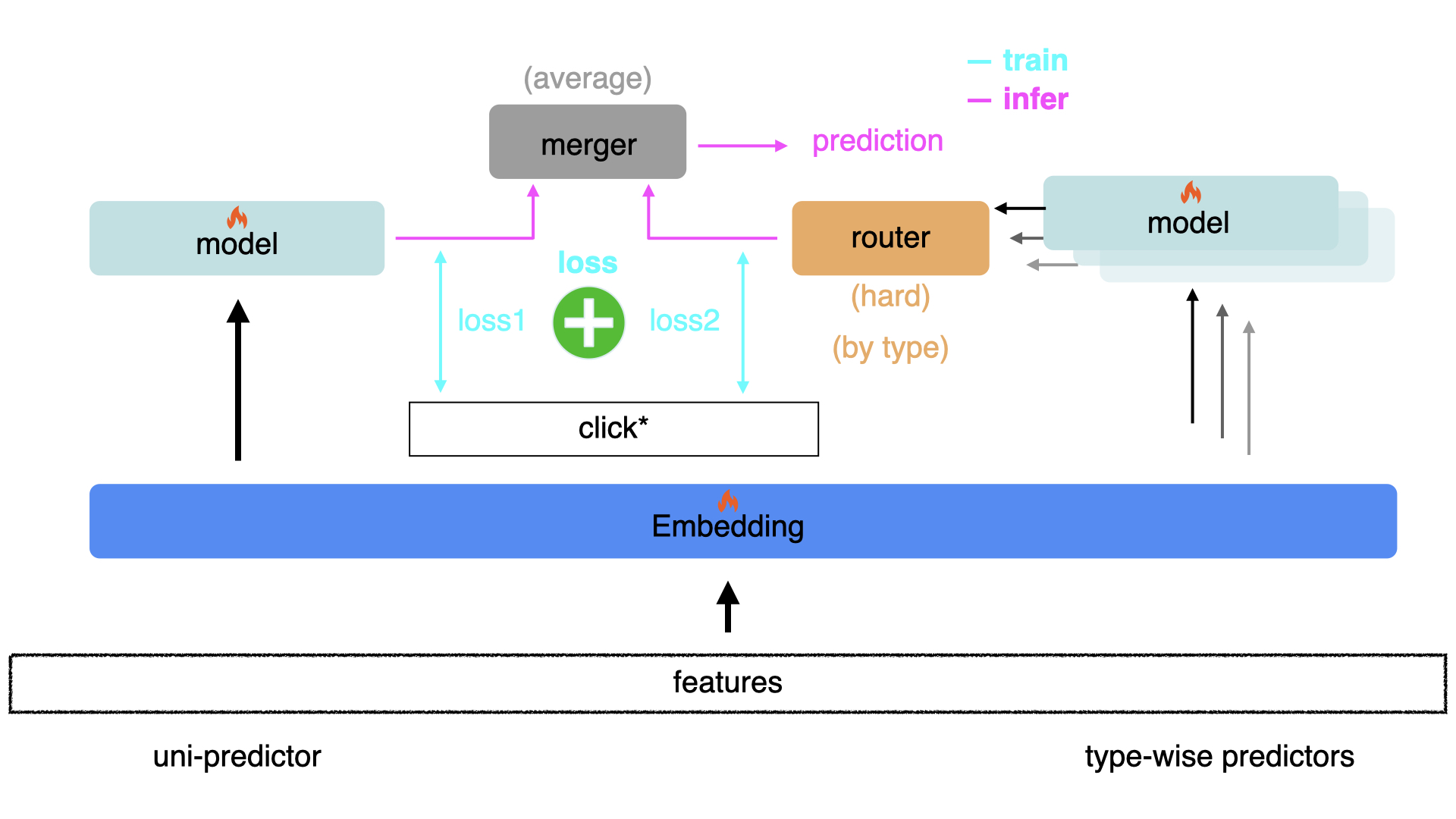}
\caption{Illustration on the ensemble method of Whole MLPs (\textbf{WM}).}
\end{figure*}

\begin{figure*}[ht]
\centering
\small
\includegraphics[width=2.0\columnwidth]{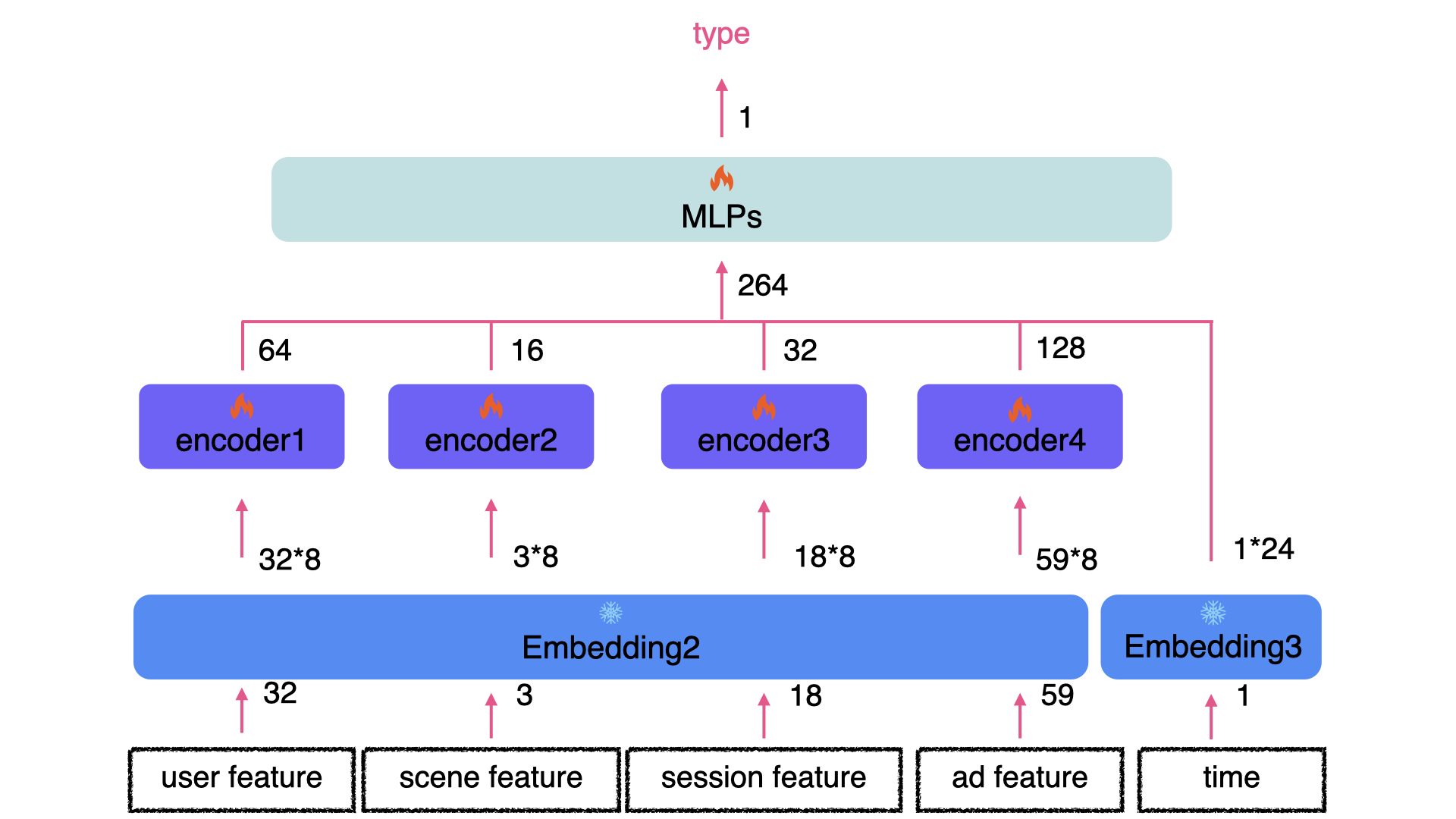}
\caption{Illustration on the structure of Auxiliary Model (\textbf{AM}).}
\end{figure*}

\begin{figure*}[ht]
\centering
\small
\includegraphics[width=2.0\columnwidth]{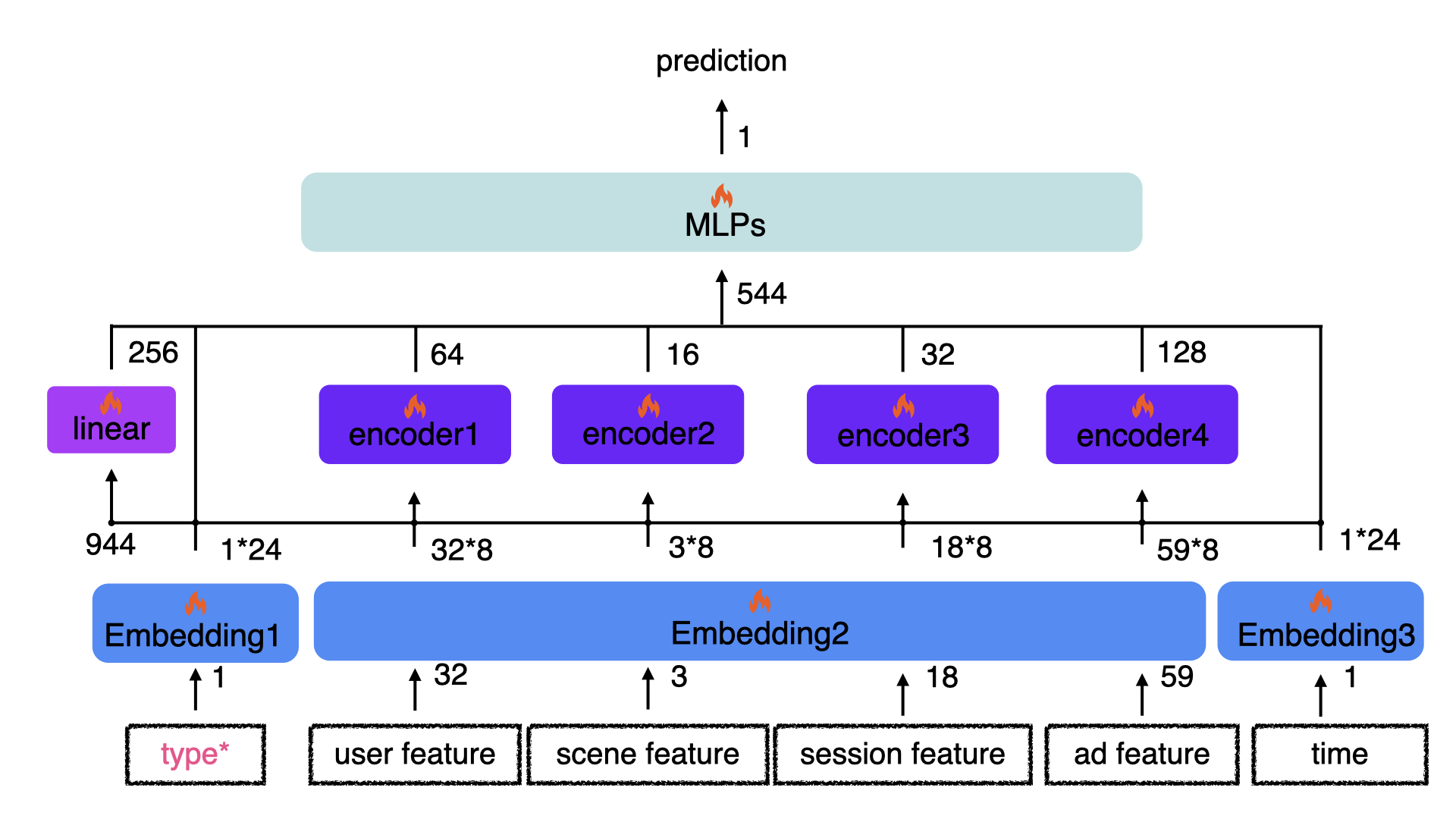}
\caption{Illustration on the structure of Hierarchical MLPs (\textbf{HM}).}
\end{figure*}

\begin{figure*}[ht]
\centering
\small
\includegraphics[width=2.0\columnwidth]{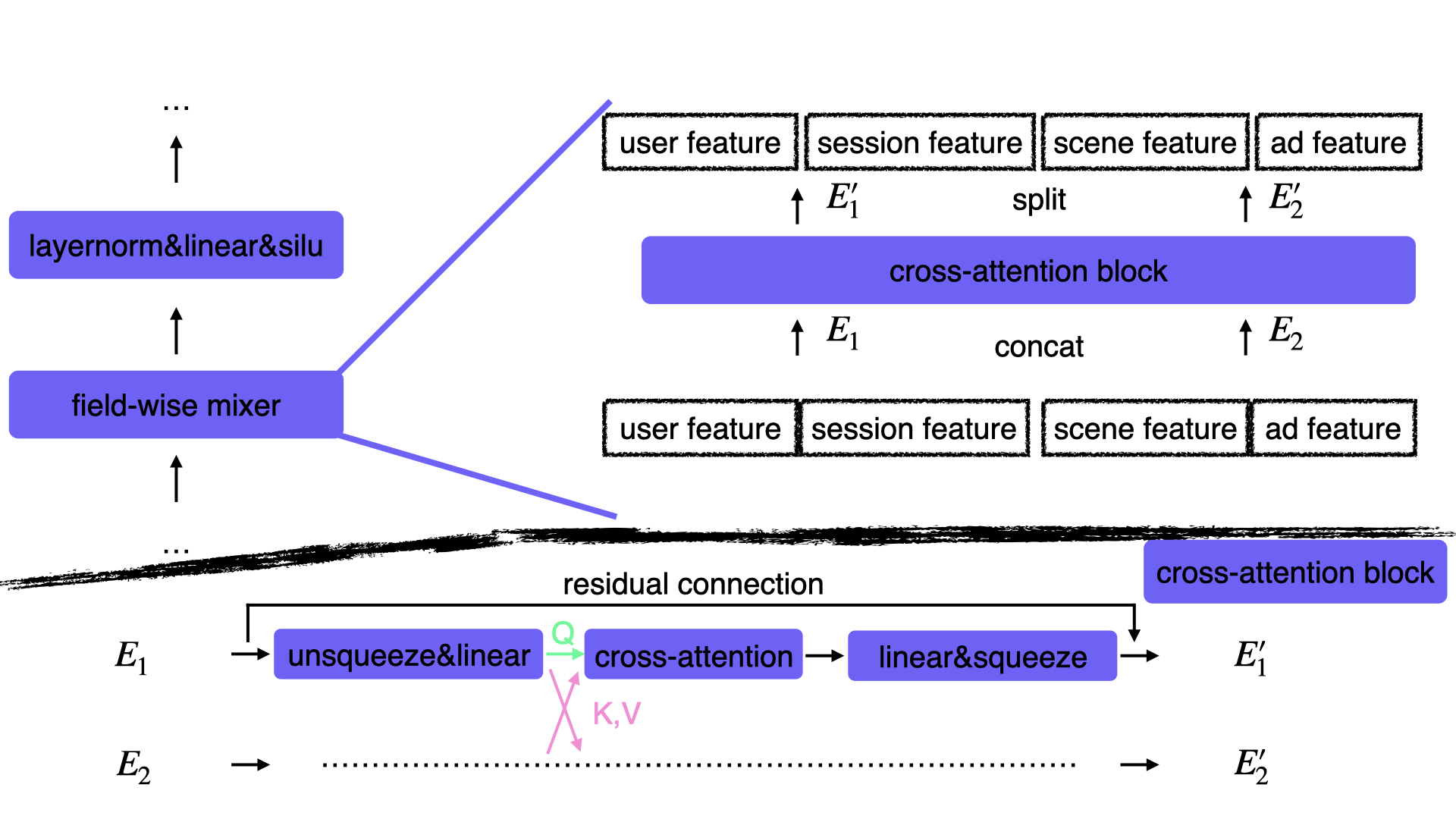}
\caption{Illustration on the feature-wise mixer of Hierarchical MLPs
with Mixers (\textbf{HMM}).}
\end{figure*}

\begin{figure*}[ht]
\centering
\small
\includegraphics[width=2.0\columnwidth]{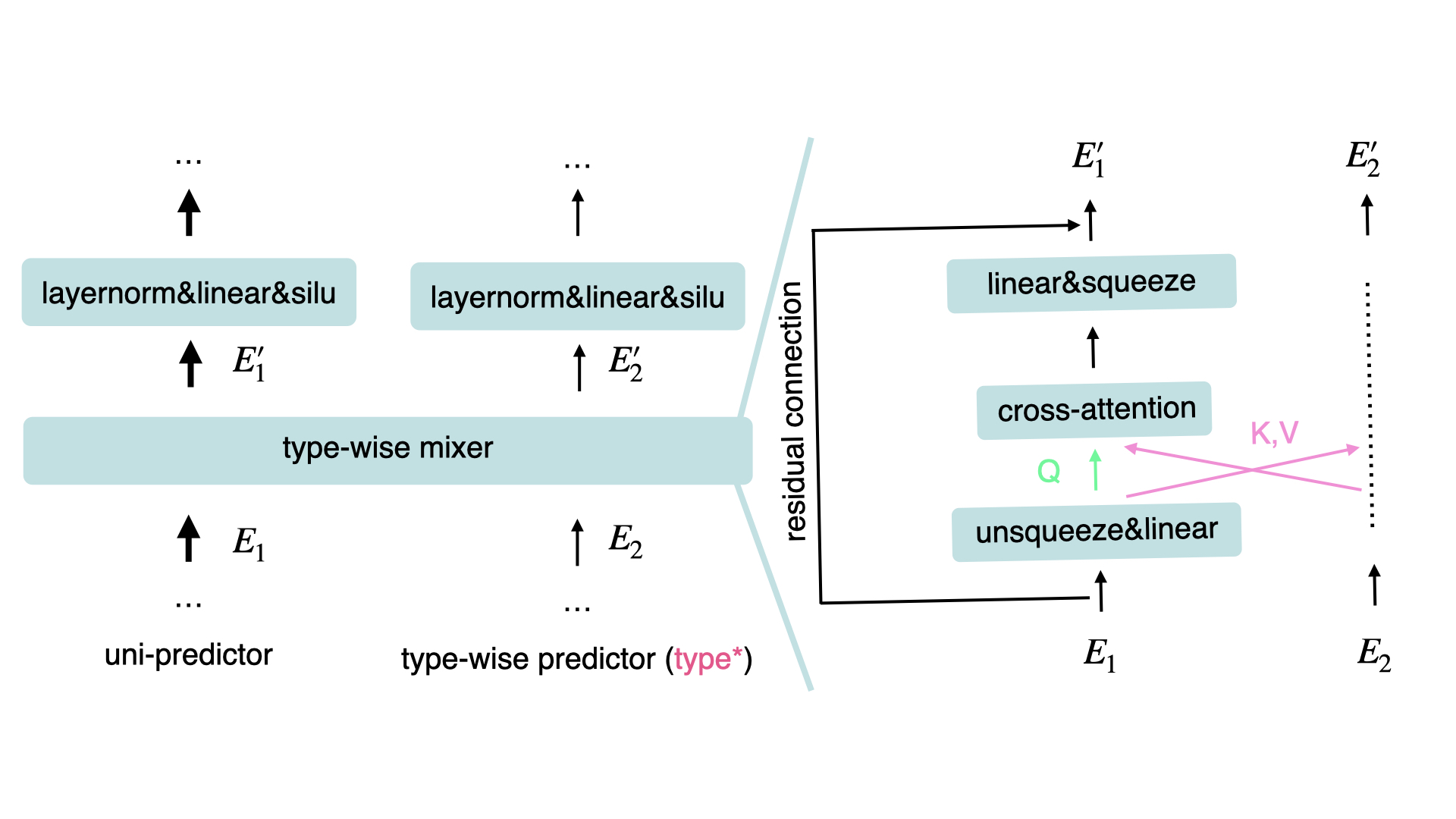}
\caption{Illustration on the type-wise mixer of Hierarchical MLPs
with Mixers (\textbf{HMM}).}
\end{figure*}

\end{document}